\documentclass{article}
\pdfoutput=1 
\usepackage[final,nonatbib]{nips_2017_edited}

\usepackage[utf8]{inputenc} 
\usepackage[T1]{fontenc}    
\usepackage{hyperref}       
\usepackage{url}            
\usepackage{booktabs}       
\usepackage{amsfonts}       
\usepackage{nicefrac}       
\usepackage{microtype}      
\usepackage{amsmath}       
\usepackage{amsthm}
\usepackage{gensymb}
\usepackage{wrapfig}
\usepackage{bm}
\usepackage{amssymb}

\usepackage{framed}
\usepackage{tikz}
\usepackage{float}
\usepackage{multirow}
\usepackage{todonotes}
\usepackage{changebar}
\usepackage{soul}
\usepackage{color}
\usepackage{authblk}
\usepackage[english]{babel}
\usepackage{algorithm}
\usepackage[noend]{algpseudocode}
\usepackage{nicefrac}       
\usepackage{times}
\usepackage{graphicx} 
\usepackage{enumitem}
\usepackage{array}
\usepackage{pbox}
\usepackage{scrextend}
\usepackage{caption}
\usepackage{subcaption}

\DeclareMathOperator*{\argmin}{arg\,min}

\newcommand\footnoteref[1]{\protected@xdef\@thefnmark{\ref{#1}}\@footnotemark}

\newtheorem{lemma}{Lemma}

\newcolumntype{L}[1]{>{\raggedright\arraybackslash}p{#1}}
\newcolumntype{C}[1]{>{\centering\arraybackslash}p{#1}}
\newcolumntype{R}[1]{>{\raggedleft\arraybackslash}p{#1}}

\newcommand{\FsNorm }[1]{\mbox{}\|#1\|_F  }

\newcommand{\setlinespacing}[1]%
           {\setlength{\baselineskip}{#1 \defbaselineskip}}

\newcommand{\sabs }[1]{|#1|}

\newcommand{\mat}[1]{{\ensuremath{\bm{\mathrm{#1}}}}}

\def\w{{\mathbf w}}

\def\x{{\mathbf x}}

\def\matA{\mat{A}}

\def\matD{\mat{D}}

\def\matM{\mat{M}}

\def\matW{\mat{W}}

\def\frac#1#2{{#1\over #2}}

\def\argmin{\mathop{\hbox{argmin}}\limits}
\def\argmax{\mathop{\hbox{argmax}}\limits}

\def\x{{\mathbf x}}

\def\floor#1{{\left\lfloor\,#1\,\right\rfloor}}

\def\dotfil{\leaders\hbox to 1.5mm{.}\hfill}
\newcounter{rmnum}
\def\RN#1{\setcounter{rmnum}{#1}\uppercase\expandafter{\romannumeral\value{rmnum}}}
\def\rn#1{\setcounter{rmnum}{#1}\expandafter{\romannumeral\value{rmnum}}}

\title{Ternary Neural Networks with Fine-Grained Quantization}

\author[1]{Naveen Mellempudi}
\author[1]{Abhisek Kundu}
\author[1]{Dheevatsa Mudigere}
\author[1]{Dipankar Das}
\author[1]{Bharat Kaul}
\author[2]{Pradeep Dubey}
\affil[1]{Parallel Computing Lab, Intel Labs, Bangalore, India}
\affil[2]{Parallel Computing Lab, Intel Labs, Santa Clara, CA}

\begin{document}

\maketitle

\begin{abstract}
We propose a novel fine-grained quantization (FGQ) method to ternarize pre-trained full precision models, while also constraining activations to 8 and 4-bits. Using this method, we demonstrate minimal loss in classification accuracy on state-of-the-art topologies without additional training. We provide an improved theoretical formulation that forms the basis for a higher quality solution using FGQ. Our method involves ternarizing the original weight tensor in groups of $N$ weights. Using $N=4$, we achieve Top-1 accuracy within $3.7\%$ and $4.2\%$ of the baseline full precision result for Resnet-101 and Resnet-50 respectively, while eliminating $75\%$ of all multiplications. These results enable a full 8/4-bit inference pipeline, with best reported accuracy using ternary weights on ImageNet dataset, with a potential of $9\times$ improvement in performance. Also, for smaller networks like AlexNet, FGQ achieves state-of-the-art results. We further study the impact of group size on both performance and accuracy. With a group size of $N=64$, we eliminate $\approx99\%$ of the multiplications; however, this introduces a noticeable drop in accuracy, which necessitates fine tuning the parameters at lower precision. We address this by fine-tuning Resnet-50 with 8-bit activations and ternary weights at $N=64$, improving the Top-1 accuracy to within $4\%$ of the full precision result with $<30\%$ additional training overhead. Our final quantized model can run on a full 8-bit compute pipeline using 2-bit weights and has the potential of up to $15\times$ improvement in performance compared to baseline full-precision models.

\end{abstract}

\section{Introduction}
Today's deep learning models achieve state-of-the-art results on a wide variety of tasks including Computer Vision, Natural Language Processing, Automatic Speech Recognition and Reinforcement Learning \cite{2016DlBook}. Mathematically, this involves solving a non-convex optimization problem with order of millions or more parameters. Solving this optimization problem - also referred to as training the neural network - is a compute-intensive process that, for current state-of-the-art networks, requires days to weeks. Once trained, the network evaluates a function on specific input data - referred to as inference. While the compute intensity for inference is much lower than that of training, owing to the fact that inference is done on a large number of input data, the total computing resources spent on inference is likely to dwarf those spent on training. The large and somewhat unique compute requirements for both deep learning training and inference operations motivate the use of customized low precision arithmetic \cite{hubara2016qnn,courbariaux2016bnn,hubara2016bnn,zhou2016dorefa,logqunat,2016twn} and specialized hardware to run these computations as efficiently as possible \cite{gupta2015lp,zhu2016ttq,venkatesh2016,finn2016,tpuGblog}. Most of these cases requires partial or full training of network in low precision.
Training at low-precision allows for the network to implicitly learn the low precision representation (along with the inherent noise); however, it introduces significant resource overheads which can be prohibitive for many resource-constrained applications, specifically those involving edge devices.

Reducing precision for both weights and activations has significant power-performance implication on system design. Low-precision not only allows increasing compute density, but also reduce pressure on the memory sub-system. Most of the current solutions are focused on compressing the model~\cite{NNFM2015,rastegariECCV16}, going as low as binary weights, which allows storing the model on the limited on-chip local memory. However, activations (input) need to be fetched from external memory or I/O-device (camera). Fetching data contributes to majority of the system power consumption. Hence reducing the size of activations is essential for more efficient utilization of the available computational resources. There have been a few solutions~\cite{hubara2016bnn,hubara2016qnn} using lower precision representation for activations, however they necessitate specialized hardware for efficient implementation. Further, with widespread adoption of deep learning across various applications, such as autonomous driving, augmented reality etc, there is an increased demand for inference tasks to be done on edge devices efficiently. To address both the aforementioned system and application requirements, there is a general trend to move towards a \emph{full lower precision inference pipeline}~\cite{tpuGblog}. This is evident from the advent of 8-bit and sub 8-bit hardware such as Google's TPU~\cite{tpuGblog}
and other main stream GPU\footnote{\label{nv8b}https://devblogs.nvidia.com/parallelforall/new-pascal-gpus-accelerate-inference-in-the-data-center/}, CPU offerings. Further, there is also software support for 8-bit inference through popular frameworks such as TensorFlow, Theano and compute libraries like NVidia's TensorRT\footnotemark[1].

In this paper, we focus on enabling a sub 8-bit inference pipeline by using ternary weights and 8/4-bit activations, with minimal or no re-training, and yet achieving near state-of-art accuracy. The rationale behind our approach is to carefully convert the full-precision weights to low-precision, such that the element-wise distance between full-precision and low-precision weights are small. Consequently, the low-precision weights remain in the neighborhood of pre-trained full-precision weights in the search space of network parameters, and we expect them to generalize in a similar manner, despite no re-training. 

We summarize our contributions below:
\begin{enumerate}
\item Based on an improved theoretical formulation, propose a novel fine-grained quantization (FGQ) method to convert pre-trained models to a ternary representation with minimal loss in test accuracy, without re-training.
%
\item With ternary weights, achieve classification accuracy (Top-1) of $73.85\%$ with 8-bit activations (2w-8a) and $70.69\%$ with 4-bit activations (2w-4a), on the ImageNet dataset\cite{imagenet_data} using a pre-trained Resnet-101 model (no re-training). To the best of our knowledge, these are the highest reported accuracies in this category on ImageNet dataset\cite{imagenet_contest}.
\item Demonstrate the general applicability of FGQ, with state-of-art results (2w-8a, 2w-4a) on smaller models such as Resnet-50 and Alexnet\cite{alexnet}. And also show the efficacy of using FGQ for (re)training at low precision.
\item Study the performance-accuracy trade-off using different group sizes. For a group of $N$ filters, we reduce the number of multiplications to one in every $N$ additions, thus significantly reducing computation complexity with a potential for up to $16\times$ improvement baseline full precision models.
\end{enumerate}

The rest of the paper is organized as follows, Section\ref{sec_litsurvey} discusses related work on ternary weights, low precision inference and contrast them with FGQ. Section\ref{sec_ternaryconversion} describes the FGQ formulation and the theoretical basis for this method. This is followed by Section\ref{sec_experiments}, which includes experimental results and related discussion. Finally, in Section\ref{sec_conclusion} we conclude with summarizing the implications of of FGQ, our results and the future research directions.

\section{Related Work}
\label{sec_litsurvey}


Deep learning inference using low-precision weights and activations is a well-researched topic. Many researchers have experimented with custom data representations to perform deep learning tasks and have shown significant benefits over the general purpose floating point representation. ~\cite{vanhoucke2011improving} have show that 8-bit dynamically scaled fixed point representation \cite{williamson1991dynamically} can be used to speed up convolution neural networks using general purpose CPU hardware by carefully choosing right data layout, compute batching and using implementation optimized for target hardware. With this they show up to $4\times$ improvement over an aggressively tuned floating point implementation. ~\cite{gupta2015lp} have done a comprehensive study on the effect of low precision fixed point computation for deep learning and have successfully trained smaller networks using 16-bit fixed point on specialized hardware. This suggests that fixed point representations are better suited for low(er) precision deep learning. There have also been recent efforts exploring 8-bit floating point representation~\cite{dettmers20158}, however such schemes have the additional overhead of reduced precision since the exponent is replicated for each value. Whereas with fixed point representation, using a single shared exponent improves capacity for precision. Typically for deep neural networks, with reduced bit-widths it is desired to preserve the numerical precision, since the loss in range can be augmented by the dynamic scaling of the shared exponent.

Commonly, low precision networks are designed to be trained from scratch,  leveraging the inherent ability of network to learn the approximations introduced by low precision computations ~\cite{NNFM2015, rastegariECCV16, hubara2016qnn, courbariaux2016bnn, hubara2016bnn, zhu2016ttq, han2015learning}. This can be prohibitive in applications which rely on using previously trained models. Such use cases are typical in many edge device deployments. To address such cases, FGQ is developed with motivation to be able to achieve state-of-art accuracies without any training and hence enabling direct use of pre-trained models. This requirement results in the quantization scheme being quite complex but making it more widely applicable and making it also easily usable for the former case - with training from scratch.

Many of the recent reduced precision work, look at the low precision only for the weights while retaining the activations in full precision~\cite{inq,venkatesh2016,han2015learning,logqunat,rastegariECCV16,NNFM2015,dettmers20158}. Using low precision also for activations is essential to realize the full power-performance benefits with using 2-bit (ternary) weights. The hardware needs to operate at the throughput that is close to the precision of the weights (i.e. $16\times$ better throughput compared to using 32-bit weights). This cannot be achieved (or would be very hard to achieve) when the activations are at full precision because streaming 32-bit activations from main memory at that rate requires much higher($16\times$) bandwidth and compute engine needs to be much wider to deliver the desired throughput. All of which will increase the area and power budget which is not desirable when designing for low-power edge devices. Hence, reducing the size of activations is essential for reducing the compute requirements at the edge. Using 8-bits and below for activations dramatically reduces the design requirements on the edge and opens up the possibility of achieving $16\times$ throughput improvement.~\cite{NNFM2015,rastegariECCV16} propose low precision networks with binary weights, while retaining the activations in full precision.~\cite{NNFM2015} use a stochastic binarization scheme, achieving state-of-art (SOTA) accuracies on smaller data-sets (MNIST, CIFAR10, SVHN).~\cite{rastegariECCV16} demonstrate near-SOTA accuracies on  the large ImageNet data-set using AlexNet topology. Further, they also demonstrate a variant with binary weights and activations, with all computations are simplified bit-count operations but with significant loss in accuracy. Lower precision for activations have also been used, \cite{hubara2016bnn} use 1-bit for both weights and activations for smaller networks. For larger Imagenet-class networks \cite{hubara2016qnn}, use 2-bit activations and binary weights showing reasonable accuracies. However, both these~\cite{hubara2016bnn,hubara2016qnn} use specialized data representation requiring custom hardware for efficient implementation. Other solutions such as~\cite{zhou2016dorefa}, employ a more tailored approach with different precision for each - weights (1-bit), activations (2-bits) and gradients (6-bits); implemented with special-purpose hardware.


~\cite{2016twn} introduces a theoretical formulation for ternary weight network using a threshold based approach (symmetric threshold $\pm \Delta$) with one scaling factor for each layer. They provide an approximation to the optimal ternary representation assuming weights follow a Gaussian distribution. However, one scaling factor per layer may not approximate the full-precision network well as the model capacity is very limited in this case. To increase model capacity ~\cite{zhu2016ttq} modify this solution to use two symmetric thresholds ($\pm\Delta$) and two scaling factors (separately for positive and negative weights). However, despite improving the accuracy this approach typically makes the inferencing inefficient by requiring multiple passes over the positive and negative values, hence increasing the bandwidth requirements. \cite{inq} have proposed a post-facto incremental quantization approach, which aims to find the optimal representation using an iterative method, constraining weights to either 0 or powers of 2, using a 5-bit representation. and re-training activations with full precision. All the aforementioned implementation require partial or full training of the network in low precision. Alternatively, \cite{logqunat} used log quantization method on pre-trained models and achieved good accuracy by tuning the bit length for each layer without re-training. 

Achieving near-SOTA accuracy on the Imagenet dataset with deeper networks \cite{resnet}, without any training in low precision (for both weights and activations) is still a challenge. Our work is an attempt to address this problem and improve over existing approaches.

\section{Ternary Conversion of Trained Network}{\label{sec_ternaryconversion}}
Our goal is to convert the full-precision trained weights $\matW$ to ternary values $\{-\alpha, 0, +\alpha\}$, $\alpha\geq 0$, without re-training. We use a threshold ($\Delta>0$) based approach similar to \cite{2016twn}: $i$-th element $\hat\matW_i = sign(\matW_i)$, if $\sabs{\matW_i}> \Delta$, and $0$ otherwise.
Then, the element-wise error is  $E(\alpha,\Delta)=\FsNorm{\matW-\alpha\hat\matW}^2$ and an optimal ternary representation $\alpha^*\hat\matW{}^*\approx \matW$ is as follows:
\begin{eqnarray}\label{eqn:twn}
\alpha^*, \Delta^* &=& \argmin_{\alpha\geq 0,\Delta>0}E(\alpha,\Delta), 
\quad 
\text{ s.t. }
\quad
\alpha \ge 0, \hat\matW_i\in \{-1,0,+1\}, i=1,2,...,n
\end{eqnarray} 
where $n$ is the size of $\matW$ ($\matW \in \mathbb{R}^n$).
We hypothesize that weights that learn different types of features may follow different distributions. Combining all the weights together represents a mixture of various distributions, and  ternarizing them using a single threshold ($\Delta$) and magnitude ($\alpha$) may not preserve the distributions of individual weights. Consequently, many weights are approximated poorly (if not totally pruned out) leading to loss of valuable information that they learn. We may not be able to compensate for this loss of information as we do not train the network in low precision. 

This motivates us to use a fine-grained quantization technique involving multiple scaling factors in order to increase model capacity that can lead to better preservation of distributions learned by filters. Moreover, we hypothesize that positive and negative weight distributions are not always symmetric around the mean, further refinement of this solution maybe possible using two separate thresholds, $\Delta_p$ and $\Delta_n >0$, for  positive and negative weights, respectively, along with a scaling factor $\alpha$ to ternarize the weights.
%


\subsection{Our Formulation}{\label{sec_formulation}}
Computing separate $\Delta$ and $\alpha$ for each weight compensates for information loss and better preserves the underlying distributions. However, such solution, while showing significant improvement in accuracy, does not reduce the number of multiplications leading to a less efficient implementation. Therefore, we seek to find a trade-off between achieving higher accuracy and reducing the total number of multiplications. We propose a fine-grained quantization approach that creates groups of weights, and ternarizes each group independently.  
Let us consider the weights represented as a vector $\matW \in \mathbb{R}^n$. We partition the set $I$ of $n$ indices into $k$ disjoint subsets, $c_1, c_2, ..., c_k$, with cardinality $\sabs{c_i}=n_i$, such that, $c_i\cap c_j = \emptyset$, $\cup_i c_i = I$, $\sum_in_i = n$. We can decompose $\matW$ into $k$ orthogonal vectors $\matW^{(i)} \in \mathbb{R}^n$, $i=1,...,k$, where $j$-th component $\matW^{(i)}_j=\matW_j$ if $j \in c_i$, otherwise $0$. Clearly, $\sum_i \matW^{(i)} = \matW$; then we ternarize each orthogonal component $\matW^{(i)}$ as $\alpha_i\hat\matW{}^{(i)}$, where components of $\hat\matW{}^{(i)}$ are in $\{-1, 0, +1\}$. Threshold-based pruning never turns $0$ to non-zero, and the following orthogonality holds.
%
$
\matW^{(i)}\perp\matW^{(j)}, 
\hat\matW{}^{(i)}\perp \hat\matW{}^{(j)}, 
\matW^{(i)}\perp\hat\matW{}^{(j)}, \text{ for } i\neq j
$. 
It follows that,  
$
(\matW^{(i)}-\alpha_i\hat\matW{}^{(i)})\perp (\matW^{(j)}-\alpha_j\hat\matW{}^{(j)})
$, for $i\neq j$, and we have
$
\FsNorm{\matW - \sum_i\alpha_i\hat\matW{}^{(i)}}^2
= 
\sum_i \FsNorm{\matW^{(i)}-\alpha_i\hat\matW{}^{(i)}}^2
$.
Then, for a given group of $k$  filters $\{\matW^{(i)}\}, i=1,...,k$, and $ \hat\matW{}^{(i)}_j\in \{-1,0,+1\}, \forall j$,
\begin{eqnarray}\label{eqn:main}
\alpha_1^*,..,\alpha_k^*,\hat\matW{}^{(1)}{}^*,..,\hat\matW{}^{(k)}{}^* 
=
{\argmin_{\alpha_i,\hat\matW{}^{(i)}
}}
\FsNorm{\matW - \sum_i\alpha_i\hat\matW{}^{(i)}}^2,
=
\sum_i
{\argmin_{\alpha_i,\hat\matW{}^{(i)}}}
\FsNorm{\matW^{(i)}-\alpha_i\hat\matW{}^{(i)}}^2
\end{eqnarray}
Therefore, we need to solve $k$ independent sub-problems. This formulation allows a better ternary approximation to the original full-precision weights, ensuring that they remain within a neighborhood of the original solution in the complex search space of parameters, despite no re-training. Consequently, we expect the full-precision solution and the ternary counterpart to generalize in a similar manner.
From model capacity point of view, we can have only three distinct values, $\{-\alpha,0,+\alpha\}$, for a ternary weight vector without such grouping. With $k$ such groups, however, we can represent $2k+1$ distinct values, thus increasing model capacity linearly with number of groups.

We can solve each sub-problem using a threshold-based approach the following way. We are given a vector of elements $\matW \in \mathbb{R}^n$, and we can use separate thresholds for positive and negative weights, along with one scaling factor $\alpha$ to ternarize $\matW$. Let
$
I_\Delta^+ = \{i: \matW_i > \Delta_p\}
$,
$
I_\Delta^- = \{i: \matW_i < -\Delta_n\}
$, for some $\Delta_p,\Delta_n >0$.
We want to solve
\begin{eqnarray}\label{eqn:asymmetry}
\alpha^*,\Delta_p^*,\Delta_n^*
= \argmin_{\alpha\geq 0,\Delta_p>0,\Delta_n>0}
\FsNorm{\matW - \alpha\hat\matW}^2,
\quad \text{s.t.}
\quad
\hat\matW_i\in \{-1,0,+1\}, i=1,...,n.
\end{eqnarray}
We have the following analytical solution.
\begin{eqnarray}\label{eqn:asymmetry_sol}
\Delta_p^*,\Delta_n^* = \argmax_{\Delta_p>0,\Delta_n>0} \frac{(\sum_{i\in I_\Delta^+}\sabs{\matW_i}+\sum_{i\in I_\Delta^-}\sabs{\matW_i})^2}{\sabs{I_\Delta^+} + \sabs{I_\Delta^-}}
, \quad 
\alpha^* = \frac{\sum_{i\in I_\Delta^+}\sabs{\matW_i}+\sum_{i\in I_\Delta^-}\sabs{\matW_i}}{\sabs{I_\Delta^+} + \sabs{I_\Delta^-}}
\end{eqnarray}
Note that for $\Delta_p=\Delta_n=\Delta$, $I_\Delta = I_\Delta^+\cup I_\Delta^-$, and (\ref{eqn:asymmetry_sol}) reproduces the formulation in (\cite{2016twn}). 
\begin{eqnarray}\label{eqn:symmetric}
\alpha^* 
= (\sum_{i\in I_\Delta}\sabs{\matW_i})/\sabs{I_\Delta}
,\quad
\Delta^* = \argmax_{\Delta>0}(\sum_{i\in I_\Delta}\sabs{\matW_i})^2/\sabs{I_\Delta}
\end{eqnarray}

The advantage of our formulation (\ref{eqn:main}) is that the smaller independent sub-problems can be solved efficiently using brute-force methods to achieve better approximation. 
However, we also explore analytical solutions to establish the theoretical veracity of our approach.
Assuming that the magnitude of the learned weights follow exponential distribution with parameter $\lambda$, we  analytically derive the optimal $\Delta^*$ from the following lemma.
\begin{lemma}\label{lemma:main}
Using above notations, 
if $\sabs{\matW_i} \sim exp(\lambda)$, then, $\Delta^*$ in (\ref{eqn:symmetric}) is  
$
\Delta^*  
\approx 1/\lambda
=\sum_i\sabs{\matW_i}/n
$
\end{lemma}

\begin{wrapfigure}{r}{0.5\textwidth}
\centering
\includegraphics[width=.5\textwidth,height=5cm,keepaspectratio]{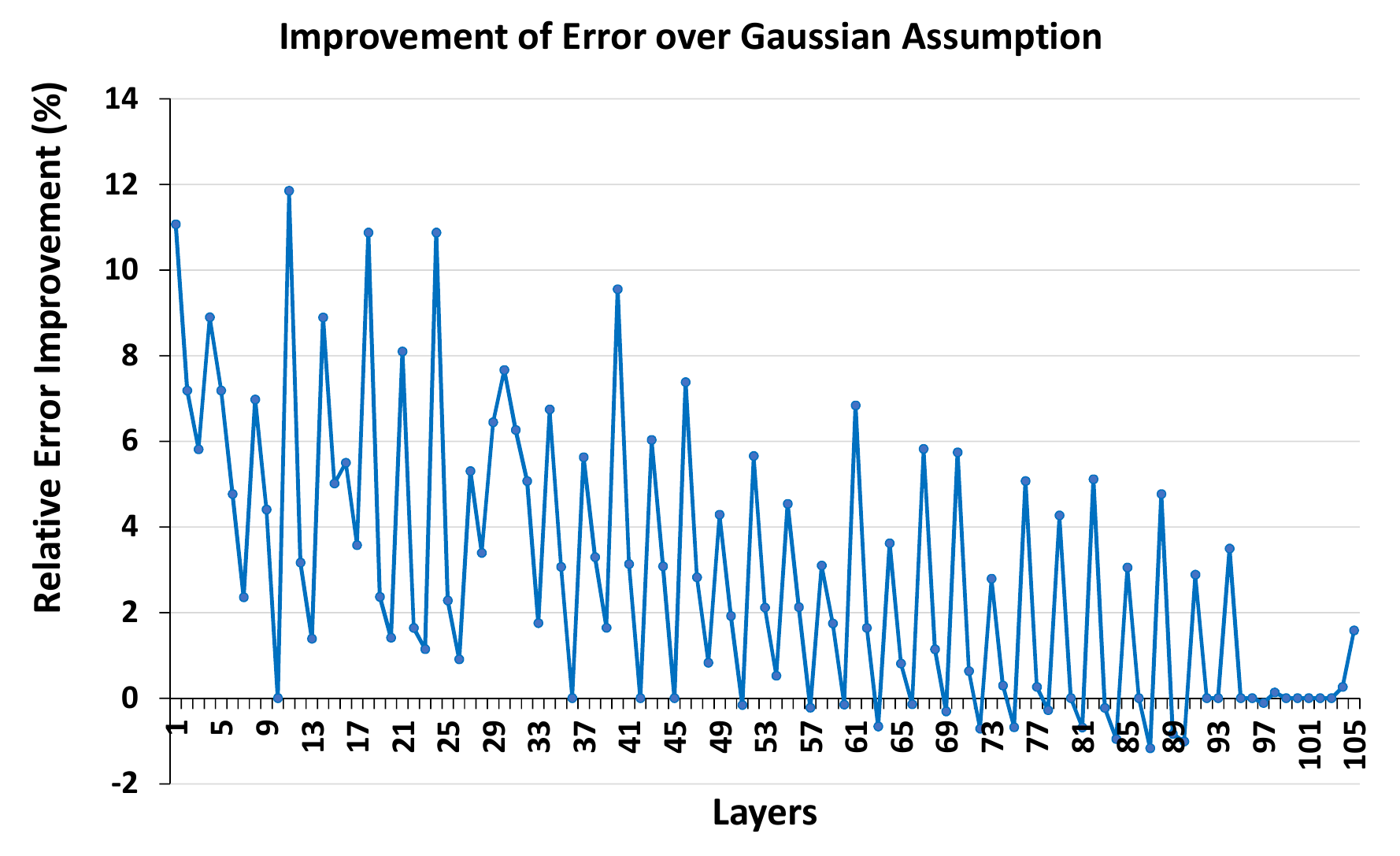}
\caption{Layer-wise improvement of theoretical ternary error (one $\alpha$ per layer) over Gaussian assumption by choosing an appropriate distribution using K-S test for ResNet-101 on Imagenet dataset.} 
\label{fig:err_improvement}
\end{wrapfigure}

From this analysis, we see the need for a higher threshold value to prune larger number of smaller elements. This is intuitive from the shape of the model distributions, which are typically heavy-tailed distributions. 
In reality, however, it may not be appropriate to use a single distribution to model the weights of all the layers of a neural network. We can apply Kolmogorov-Smirnov (K-S) test as a goodness-of-fit measure to identify an appropriate reference distribution (here we choose  between Gaussian and exponential), and find  $\Delta^*$ accordingly. We approximate a heavy-tailed distribution by an exponential one by pruning out some of the smaller elements. This gives us an exponential approximation with smaller $\lambda$. Further, we can use maximum likelihood functions to estimate the parameters of such distributions. For Gaussian $\mathcal{N}(0,\sigma)$, estimated $\hat\sigma = \sqrt{\sum_{i=1}^n\sabs{\matW_i}^2/n}=rms(\matW)$, and for exponential case, estimated parameter $\hat\lambda = \sum_{i=1}^n\sabs{\matW_i}/n$.
Based on such refined analysis, we observe significant improvement in the theoretical ternary error over Gaussian assumption of \cite{2016twn} (Figure \ref{fig:err_improvement}). It is interesting to observe that for earlier convolution layers of ResNet-101 trained on ImageNet, the magnitude of the weights follow exponential distribution, and later layer weights are Gaussian. 

\subsection{Weight Grouping }{\label{sec_weightgroups}}
Our method (\ref{eqn:main}) is agnostic to how the (full-precision) weights are grouped
, but leverages that consequence of grouping - which allows for solving these as independent sub-problems more efficiently. The specifics of the grouping mechanism and memory layout used for accessing these groups of weights is an independent problem to explore. The primary objective of grouping is to minimize the dynamic range within each group and split the weights in such a way that the smaller groups have a uniform distribution. This helps in reducing the complexity of finding an optimal solution ($\alpha$) for each independent sub-problem using either analytical or brute-force techniques. 

However, to realize the full performance potential of ternarization, it is essential to ensure that the grouping mechanism itself does not introduce a significant overhead.
%
%
Similarity based clustering algorithms such as K-means, despite being better at finding optimal grouping of weights that may even lead to better accuracy, are not friendly for efficient implementations (in both software hardware), because of the random grouping of elements from non-contiguous memory locations. This leads to irregular memory accesses with longer latencies, to gather arbitrarily grouped weights that use a common $\alpha$, for a partial accumulation of the output.
%

\begin{wrapfigure}{l}{0.5\textwidth}
\centering
\includegraphics[width=0.5\textwidth] {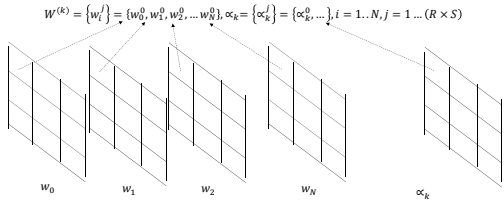}
\caption{Static Grouping: $(R\times S)$ sub-groups of $n$ elements from contiguous filters along $C$ dimension, $\alpha_k$ consists of scaling factors.}
\label{fig:clusterdiagrm}
\end{wrapfigure} 

Based on our empirical observations, we conclude that using static groups of weights that are partitioned along input channels $C$ achieves best accuracy. The same element from multiple filters along $C$ have significantly less variance, since they correspond to similar input features. Hence grouping such elements results in reduced dynamic range within the group. Such a grouping also easily lends itself for efficient implementation using both existing hardware and in software with using $[K][C/N][R*S][N]$ layout for the weight tensor, where groups of $N$ elements are accessed from contiguous memory locations. 
Since the elements along $C$ accumulate to the same output feature, this layout is also amenable to efficient vectorization along $K$. 
Figure  \ref{fig:clusterdiagrm} shows an example of this grouping scheme applied to $3\times 3$ filters. Each group of $N$ $3\times 3$ ternary filters, has $3\times 3$ scaling factors ($\alpha$) corresponding to each element of the filter. 

\section{Experimental Results}
{\label{sec_experiments}}

\begin{wrapfigure}{r}{0.5\textwidth}
\centering
\centerline{\includegraphics[width=0.5\textwidth,height=5cm,keepaspectratio]{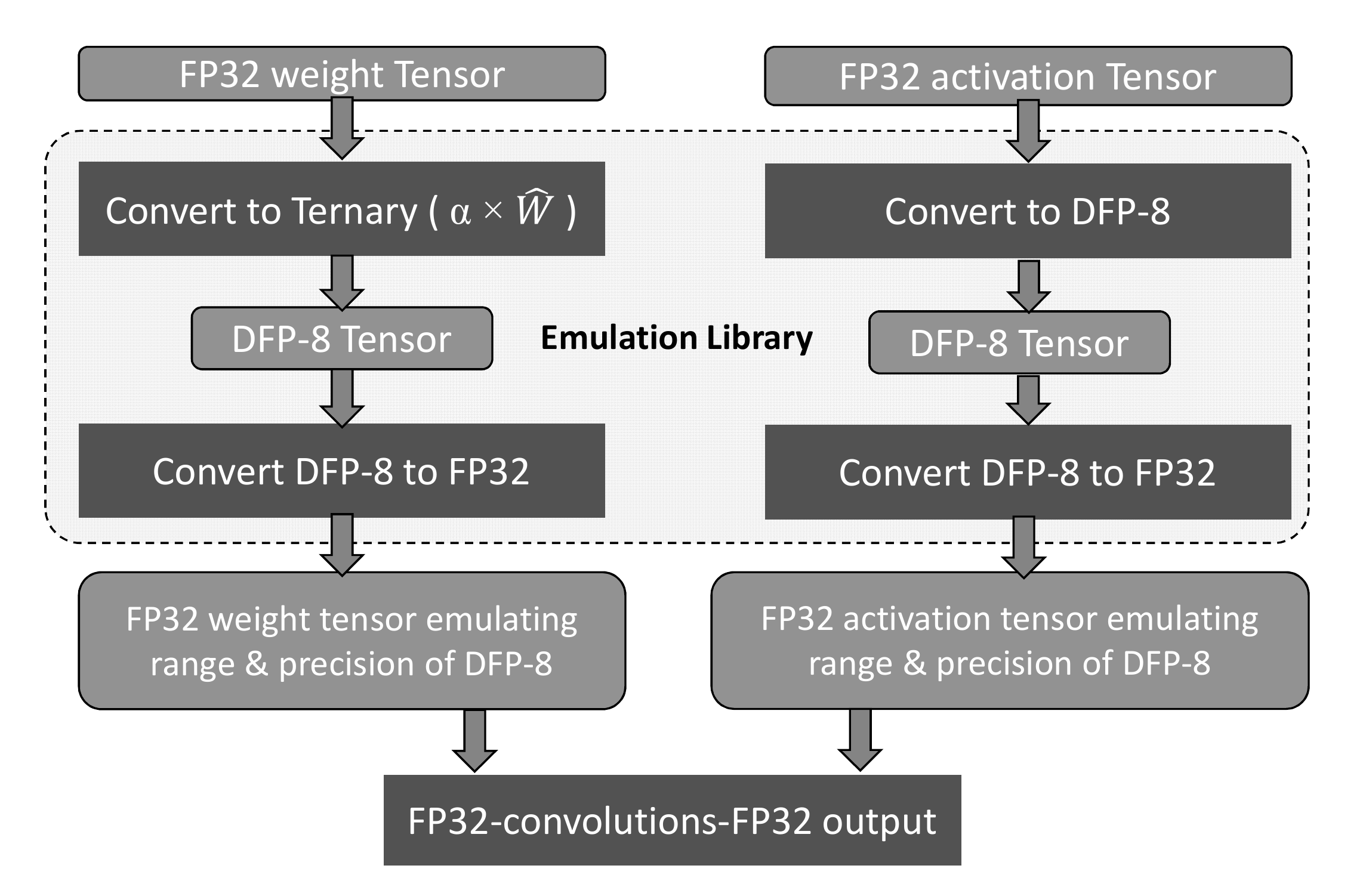}}
\caption{Schematic describing our low precision experimental setup in Caffe, to emulate fine-grained quantization (FGQ) with ternary weights and 8-bit activations}
\label{fig:exp_setup}
\end{wrapfigure}

For experimental results, we focused on Resnet-50 and Resnet-101\cite{resnet} using ILSVRC-2012\cite{imagenet_data} dataset, to demonstrate the efficacy of our method on large, sophisticated models using 2-bit weights and 8-bit activations (2w-8a). We extended our study by applying FGQ on activations to help further reduce the precision of activations to 4-bits (2w-4a) and show results comparable with 8-bit activations for all the tested networks. Further, towards establishing the broader applicability of FGQ we demonstrate state-of-the-art accuracy also for Alexnet\cite{alexnet}.

Our setup consists of a modified version of Caffe\cite{caffe} that emulates low-precision dynamic fixed point (DFP\footnote{Please not that fixed point and dynamic fixed point are used interchangeably}) computations described in Fig.~\ref{fig:exp_setup}. We use 32-bit accumulator for all our low precision computations to minimize the chances of an overflow. We split the pre-trained weights into groups of $N$ elements using the mechanism described in section\ref{sec_weightgroups}, and use brute-force technique to compute the floating point values of the threshold ($\Delta$) and scaling-factor ($\alpha$) for each group. The scaling factors are then quantized to a 8/4-bit fixed point and the weights are stored in the memory format described in\ref{sec_weightgroups}. The activations are quantized to 8/4-bits before performing convolution operation and the 32-bit outputs are down converted to 8-bit and appropriately rounded before they are passed to the next layer.

Our experiments indicate that it is essential to use higher precision of  the first layer (\textbf{8w-8a}), to minimize the accumulation of quantization loss. We also observe that using pre-trained parameters in batch normalization layers leads to a loss of accuracy due to shift in variance introduced by the quantization. We prevent this loss by recomputing batch normalization parameters during the inference phase to compensate for the shift in variance. 

We explored the accuracy-performance trade-off using different group sizes of $N$, our experiments show that FGQ with a group size of N=4 (\textbf{FGQ-N4}) achieves highest accuracy with no re-training and a potential \boldmath{$9\times$} performance benefit. FGQ-N4 applied to a pre-trained Resnet-101 model with \textbf{2w-8a} achieves Top-1 accuracy of \boldmath{$73.9\%$}, which is within \boldmath{$4\%$} of the full-precision results. With  activations reduced to 4-bits (\textbf{2w-4a}), the Top-1 accuracy drops only marginally to \boldmath{$70.7\%$}. FGQ-N4 performs equally well on Resnet-50, achieving with Top-1 accuracy of \boldmath{$70.8\%$} with \textbf{2w-8a} which is \boldmath{$~4.2\%$} off from full-precision result, and \boldmath{$68.4\%$} with \textbf{2w-4a}. \emph{To the best of our knowledge, these are the highest reported accuracies using 2w-8a and 2w-4a on Imagenet dataset\cite{imagenet_data} using state-of-the-art networks.} 

To understand the general applicability of our method to a wider range of networks, we apply FGQ to the smaller Alexnet\cite{alexnet} model. FGQ-N4 applied to a pre-trained Alexnet model, achieves $49\%$ Top-1 accuracy with \textbf{2w-8a} without any re-training, this is $~8\%$ away from the baseline full-precision result( $56.8\%$). With \textbf{2w-4a} we do not see any further reduction in accuracy. There are no previously published results that can be directly compared to FGQ, which perform quantizaion on pre-trained models and work with end-to-end low-precision. Hence, we compare with \cite{venkatesh2016,inq, zhou2016dorefa}, which are the closest in terms of the networks used and/or the target precision. Our Alexnet result using FGQ-N is comparable to previously published result\cite{zhou2016dorefa} which is $6\%$ away from the baseline using 1w-4a while also employing training in low-precision with full precision gradients. Table~\ref{resultstable} has a comparison with previous reported results from\cite{inq} using $5$-bit weights and \cite{venkatesh2016} using ternary weights. While they report slightly better absolute numbers, our numbers are relatively better because both these results use full-precision activations and train the network in low precision to achieve those numbers. While without any low precision training and reduced precision for activation, results with FGQ is still competitive with other similar (aforementioned) results.\footnote{\label{fnote1}It should be noted that both these works use Resnet-50 with slight variations and hence have slightly different baseline accuracies. For \cite{inq} the baseline full precision a Top-1 accuracy is $73.22\%$, for \cite{venkatesh2016} it is $76\%$ and for \cite{zhou2016dorefa} it is $55.9\%$} With additional low precision training  with FGQ we are able significantly improve accuracy and get closer to state-of-art full precision results, as outlined in the next section along with associated performance implications.

\begin{table}[t]
\centering
\small {
\caption{Classification accuracy (Top-1) on ImageNet dataset achieved using FGQ with N=4 without any re-training. Results on Resnet-101, Resnet-50 and Alexnet using 2w-8a and 2w-4a compared with best published results}
\label{resultstable}
\vskip 0.15in
\begin{tabular}{C{1.5cm}|C{1.5cm}|C{1.7cm}C{1.7cm}|C{1.5cm}C{1.5cm}C{2cm}}
\toprule
\multirow{2}{*}{Networks} & \multirow{2}{*}{Our Baseline} & FGQ-N4 2w-8a & FGQ-N4 2w-4a & INQ 5w-32a \tiny{\cite{inq}} & dLAC 2w-32a \tiny{\cite{venkatesh2016}} & DoReFa 1w-4a-32g \tiny{\cite{zhou2016dorefa}} \\
\cmidrule{3-7}
\multirow{2}{*}{} & \multirow{2}{*}{} & \multicolumn{2}{C{3.4cm}|}{\footnotesize{no low precision training}} & \multicolumn{3}{C{5cm}}{\footnotesize{with low precision re-training}}\\
\midrule
Resnet-101 	& \textit{77.50\%}	& 73.85\%	& 70.69\% &	 - 		 &	-   	& -	\\
Resnet-50 	& \textit{75.05\%}  	& 70.76\%	& 68.38\% &	 74.81\%\footref{fnote1} &	73.85\%\footref{fnote1}	& -	\\
AlexNet     & \textit{56.83\%}   & 49.04\%	& 49.00\% &	 -		 &	- 		& 50.3\%\footref{fnote1}\\
\bottomrule
\end{tabular}
}
\end{table}
 
\subsection{Discussion}
In order to realize the full performance potential of ternary networks, the inference platform needs to operate at the throughput that is close to the precision of the weights. This would increase the amount of memory bandwidth required to stream activations by $16\times$ and a compute engine that is much wider to deliver the desired compute throughput. Building such solution around full-precision activations would be prohibitive in terms of areas and power requirements, whereas it is more amenable to build such solution when the activations are 8 or 4-bits.    

\begin{figure*}[t!]
    \centering
    \begin{subfigure}[t]{0.5\textwidth}
        \centering
        \includegraphics[width=\columnwidth]{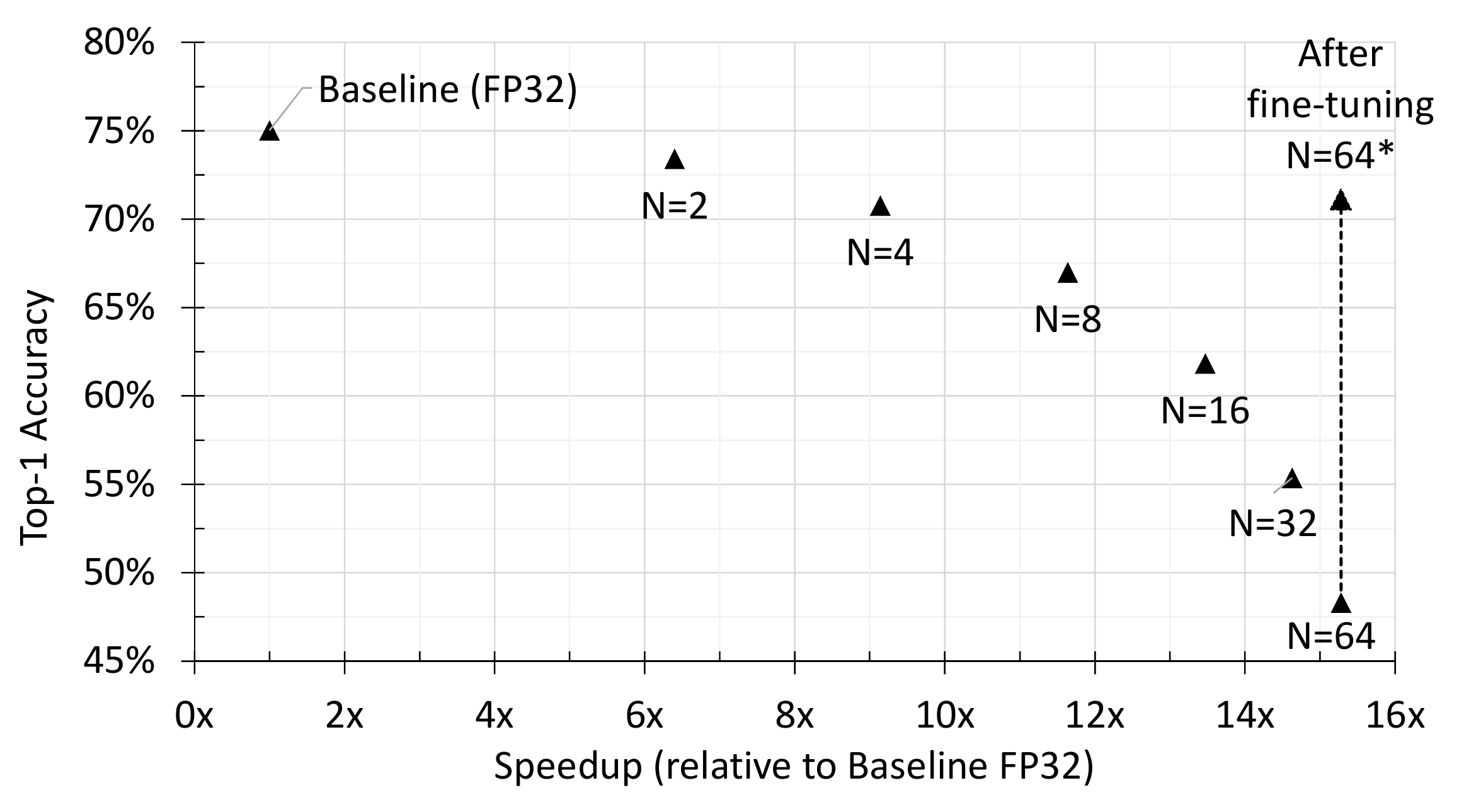}
        \caption{}
        \label{fig:perfmodel_a}
    \end{subfigure}%
    ~ 
    \begin{subfigure}[t]{0.5\textwidth}
        \centering
        \includegraphics[width=\columnwidth]{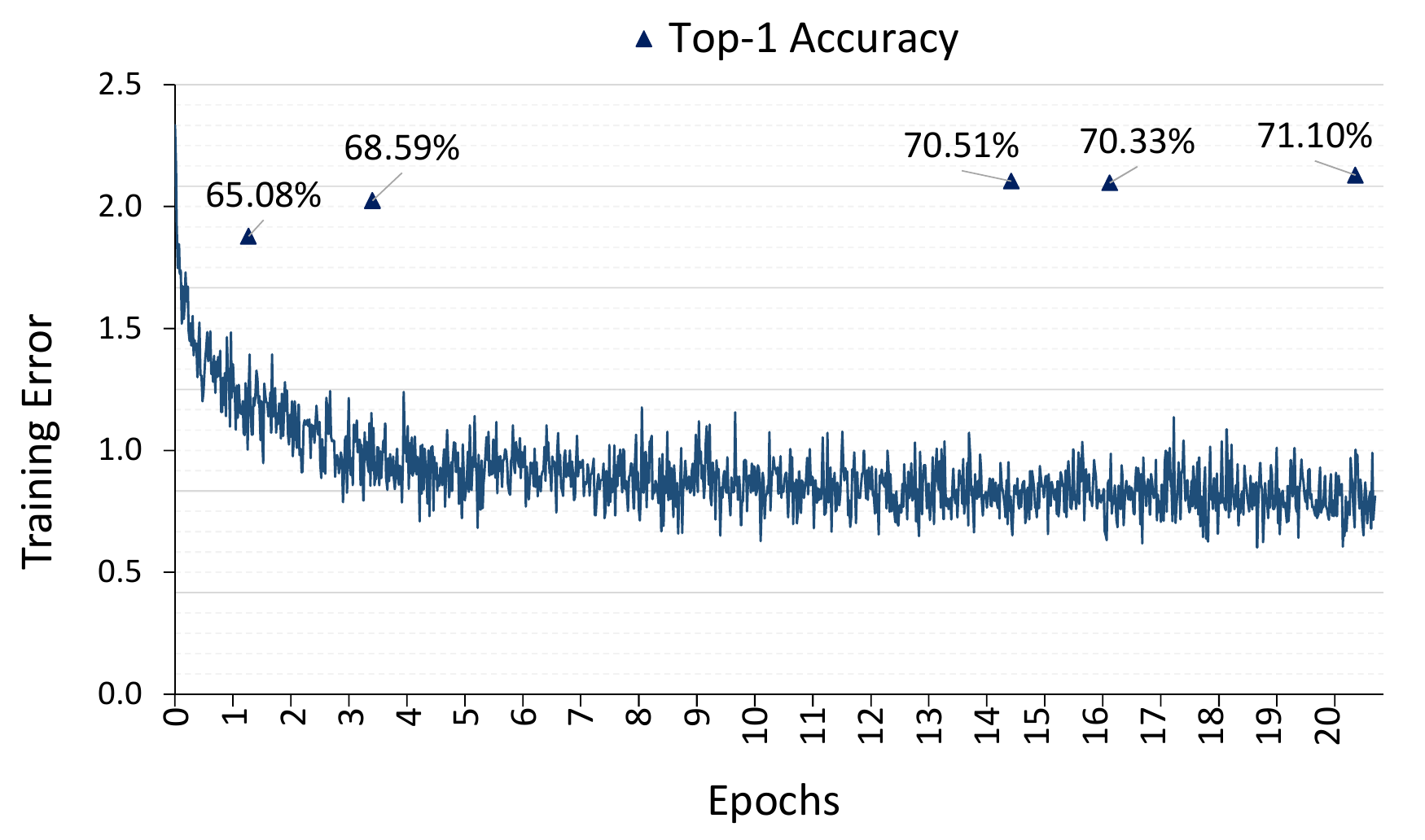}
        \caption{}
        \label{fig:perfmodel_b}
    \end{subfigure}
    \caption{a). Performance vs accuracy trade-off for Resnet-50 for a FGQ based 8-bit inference design, b). Fine-tuning Resnet-50, with pre-initialized weights on Imagenet dataset.}
\end{figure*}

Figure\ref{fig:perfmodel_a} shows the performance Vs accuracy trade-off for Resnet-50\cite{marcelbatchnorm} for a FGQ based 8-bit inference design. Our model projects the lower bound of the performance potential based on the percentage of FMA operations that can be converted into ternary accumulations at each group size $N$. In the ideal case, where $N$ is equal to the total number of weights in the layer, the best case performance potential is $16\times$ compared to the baseline full-precision performance. For a group size of N=4, $75\%$ of all FMA operations can be performed in ternary, Using slightly larger sub-groups of $\textbf{N=8}$ we can replace $87.5\%$ of FMA operations with ternary while losing an additional $3.7\%$ Top-1 accuracy. At group size $N=64$, $\approx99\%$ of all FMA operations can be replaced by ternary accumulations, resulting in \textbf{$~15\times$} potential improvement in performance. But the performance comes at a cost of significant drop in accuracy. Using larger groups of weights results in a poor ternary approximation to the full-precision model. Consequently, the ternary solution moves away from the full-precision local optima and display different generalization behavior. 


We have trained low-precision (\textbf{2w-8a}) ResNet-50\cite{marcelbatchnorm} at group size \textbf{N=64} on ImageNet\cite{imagenet_data} dataset to recover the accuracy lost because of ternarization. We initialized the network with a pre-trained full-precision model and fine-tuned the parameters of the low-precision network. 
We reduced the learning rate to an order of 1e-4 to avoid exploding gradients, retaining all the other hyper parameters from full-precision training and performed gradient updates in full precision. After training for 20-epochs, we recover most of the lost accuracy and achieved \textbf{$71.1\%$} Top-1 and $90.01\%$ Top-5 bringing it to within \textbf{$4\%$} of the full-precision baseline accuracy. Figure\ref{fig:perfmodel_b} shows the reduction of training error and improvements in validation accuracy. 



\section {Conclusion}
\label{sec_conclusion}
We propose a fine-grained ternarization method which exploits local correlations in dynamic range of the parameters to minimize the impact of quantization on overall accuracy. We demonstrate near SOTA accuracy on Imagenet data-set using pre-trained models with quantized networks without re-training. Using ternary weights on Resnet-101 and Resnet-50 with 8-bit activations our results are within $\approx4\%$ from the full precision $(FP32)$ accuracy. Using 4-bit activations we see a further drop of $\approx3\%$ in accuracy. To the best of our knowledge, these are the highest reported accuracies using ternary weights and low-precision activations.

Our weight grouping based approach allows us to obtain solutions that can be tailored for specific hardware, as well as, can be used on general purpose hardware, based on the accuracy and performance requirements. Smaller group sizes with N=4 achieve best accuracy, and use $75\%$ of the computations ternary operations (simple 8-bit additions) and this is better suited for implementation on specialized hardware. Larger group sizes are more suitable for current general purpose hardware, with a larger portion of computations as low precision operations ($\approx99\%$ for N=64),  although this comes with the cost of reduced accuracy. This gap may be bridged with additional low precision training as shown in Section\ref{sec_experiments}. Our final quantized model can be efficiently run on full 8-bit compute pipeline, thus offering a potential $16X$ performance-power benefit.

We continue to actively work on closing the current accuracy gap, exploring both low precision (re)training and extensions to the FGQ method itself. Also we are looking into a more theoretical exploration to better understand the formal relationship between the weight  grouping and final accuracy, with an attempt to establish realistic bounds for given network-performance-accuracy requirements.


\newpage
\bibliographystyle{plain}
\bibliography{references}

\begin{thebibliography}{10}

\bibitem{2016DlBook}
Yoshua Bengio, Ian Goodfellow, and Aaron Courville.
\newblock Deep learning.
\newblock {B}ook in preparation for MIT Press, 2016.

\bibitem{courbariaux2016bnn}
Matthieu Courbariaux, Itay Hubara, Daniel Soudry, Ran El-Yaniv, and Yoshua
  Bengio.
\newblock Binarized neural networks: Training deep neural networks with weights
  and activations constrained to+ 1 or-1.
\newblock {\em arXiv preprint arXiv:1602.02830}, 2016.

\bibitem{imagenet_data}
Jia Deng, Wei Dong, Richard Socher, Li-Jia Li, Kai Li, and Li~Fei-Fei.
\newblock Imagenet: A large-scale hierarchical image database.
\newblock In {\em Computer Vision and Pattern Recognition, 2009. CVPR 2009.
  IEEE Conference on}, pages 248--255. IEEE, 2009.

\bibitem{dettmers20158}
Tim Dettmers.
\newblock 8-bit approximations for parallelism in deep learning.
\newblock {\em arXiv preprint arXiv:1511.04561}, 2015.

\bibitem{gupta2015lp}
Suyog Gupta, Ankur Agrawal, Kailash Gopalakrishnan, and Pritish Narayanan.
\newblock Deep learning with limited numerical precision.
\newblock In {\em ICML}, pages 1737--1746, 2015.

\bibitem{han2015learning}
Song Han, Jeff Pool, John Tran, and William Dally.
\newblock Learning both weights and connections for efficient neural network.
\newblock In {\em Advances in Neural Information Processing Systems}, pages
  1135--1143, 2015.

\bibitem{resnet}
Kaiming He, Xiangyu Zhang, Shaoqing Ren, and Jian Sun.
\newblock Deep residual learning for image recognition.
\newblock In {\em Proceedings of the IEEE Conference on Computer Vision and
  Pattern Recognition}, pages 770--778, 2016.

\bibitem{hubara2016bnn}
Itay Hubara, Matthieu Courbariaux, Daniel Soudry, Ran El-Yaniv, and Yoshua
  Bengio.
\newblock Binarized neural networks.
\newblock In {\em Advances in Neural Information Processing Systems}, pages
  4107--4115, 2016.

\bibitem{hubara2016qnn}
Itay Hubara, Matthieu Courbariaux, Daniel Soudry, Ran El-Yaniv, and Yoshua
  Bengio.
\newblock Quantized neural networks: Training neural networks with low
  precision weights and activations.
\newblock {\em arXiv preprint arXiv:1609.07061}, 2016.

\bibitem{caffe}
Yangqing Jia, Evan Shelhamer, Jeff Donahue, Sergey Karayev, Jonathan Long, Ross
  Girshick, Sergio Guadarrama, and Trevor Darrell.
\newblock Caffe: Convolutional architecture for fast feature embedding.
\newblock {\em arXiv preprint arXiv:1408.5093}, 2014.

\bibitem{tpuGblog}
N~Jouppi.
\newblock Google supercharges machine learning tasks with tpu custom chip.
\newblock {\em Google Blog, May}, 18, 2016.

\bibitem{alexnet}
Alex Krizhevsky, Ilya Sutskever, and Geoffrey~E Hinton.
\newblock Imagenet classification with deep convolutional neural networks.
\newblock In {\em Advances in neural information processing systems}, pages
  1097--1105, 2012.

\bibitem{2016twn}
Fengfu Li, Bo~Zhang, and Bin Liu.
\newblock Ternary weight networks.
\newblock {\em arXiv preprint arXiv:1605.04711}, 2016.

\bibitem{NNFM2015}
Zhouhan Lin, Matthieu Courbariaux, Roland Memisevic, and Yoshua Bengio.
\newblock Neural networks with few multiplications.
\newblock {\em arXiv preprint arXiv:1609.07061}, 2016.

\bibitem{logqunat}
Daisuke Miyashita, Edward~H Lee, and Boris Murmann.
\newblock Convolutional neural networks using logarithmic data representation.
\newblock {\em arXiv preprint arXiv:1603.01025}, 2016.

\bibitem{rastegariECCV16}
Mohammad Rastegari, Vicente Ordonez, Joseph Redmon, and Ali Farhadi.
\newblock Xnor-net: Imagenet classification using binary convolutional neural
  networks.
\newblock In {\em ECCV}, 2016.

\bibitem{imagenet_contest}
Olga Russakovsky, Jia Deng, Hao Su, Jonathan Krause, Sanjeev Satheesh, Sean Ma,
  Zhiheng Huang, Andrej Karpathy, Aditya Khosla, Michael Bernstein, et~al.
\newblock Imagenet large scale visual recognition challenge.
\newblock {\em International Journal of Computer Vision}, 115(3):211--252,
  2015.

\bibitem{marcelbatchnorm}
Marcel Simon, Erik Rodner, and Joachim Denzler.
\newblock Imagenet pre-trained models with batch normalization.
\newblock {\em arXiv preprint arXiv:1612.01452v2}, 2016.

\bibitem{finn2016}
Yaman Umuroglu, Nicholas~J Fraser, Giulio Gambardella, Michaela Blott, Philip
  Leong, Magnus Jahre, and Kees Vissers.
\newblock Finn: A framework for fast, scalable binarized neural network
  inference.
\newblock {\em arXiv preprint arXiv:1612.07119}, 2016.

\bibitem{vanhoucke2011improving}
Vincent Vanhoucke, Andrew Senior, and Mark~Z Mao.
\newblock Improving the speed of neural networks on cpus.
\newblock In {\em Proc. Deep Learning and Unsupervised Feature Learning NIPS
  Workshop}, volume~1, page~4. Citeseer, 2011.

\bibitem{venkatesh2016}
Ganesh Venkatesh, Eriko Nurvitadhi, and Debbie Marr.
\newblock Accelerating deep convolutional networks using low-precision and
  sparsity.
\newblock {\em arXiv preprint arXiv:1610.00324}, 2016.

\bibitem{williamson1991dynamically}
Darrell Williamson.
\newblock Dynamically scaled fixed point arithmetic.
\newblock In {\em Communications, Computers and Signal Processing, 1991., IEEE
  Pacific Rim Conference on}, pages 315--318. IEEE, 1991.

\bibitem{inq}
Aojun Zhou, Anbang Yao, Yiwen Guo, Lin Xu, and Yurong Chen.
\newblock Incremental network quantization: Towards lossless cnns with
  low-precision weights.
\newblock {\em poster at International Conference on Learning Representations},
  2017.

\bibitem{zhou2016dorefa}
Shuchang Zhou, Yuxin Wu, Zekun Ni, Xinyu Zhou, He~Wen, and Yuheng Zou.
\newblock Dorefa-net: Training low bitwidth convolutional neural networks with
  low bitwidth gradients.
\newblock {\em arXiv preprint arXiv:1606.06160}, 2016.

\bibitem{zhu2016ttq}
Chenzhuo Zhu, Song Han, Huizi Mao, and William~J Dally.
\newblock Trained ternary quantization.
\newblock {\em arXiv preprint arXiv:1612.01064}, 2016.

\end{thebibliography}
\newpage
\section*{Appendix}
\subsection{Proof of Lemma \ref{lemma:main}
}
Let $n$ denote the number of elements. Let $f(x)=\lambda e^{-\lambda x}$ be the pdf of exponential distribution with parameter $\lambda>0$, and $F(x) = 1-e^{-\lambda x}$ be the cdf. Then,
$$
\sabs{I_\Delta} \approx n \int_{x>\Delta} f(x)dx 
= n(1-F(\Delta))
= n e^{-\lambda\Delta}
$$
Furthermore,
\begin{eqnarray*}
\sum_{i\in I_\Delta}\sabs{W_i} \approx n \int_{x>\Delta} x f(x)dx
= n \int_{x>\Delta} (\lambda e^{-\lambda x})x dx
\approx
\frac{n}{\lambda}\left(\lambda\Delta+1 \right)e^{-\lambda \Delta}
\end{eqnarray*}
Then,
\begin{eqnarray*}
G(\Delta) &=& \frac{(\sum_{i\in I\Delta}\sabs{W_i})^2}{\sabs{I_\Delta}}
= \frac{n}{\lambda^2}(1+\lambda\Delta)^2e^{-\lambda\Delta}
\\
G'(\Delta) &=& \frac{n}{\lambda^2}(2\lambda(1+\lambda\Delta)e^{-\lambda\Delta}-\lambda(1+\lambda\Delta)^2e^{-\lambda\Delta})
\\
G'(\Delta) &=& 0 \Rightarrow \Delta = \frac{1}{\lambda}
\\
G''(\Delta)\Big|_{\Delta=1/\lambda} &<& 0 \quad \text{(maxima)}
\end{eqnarray*}
Therefore,
$$
\Delta^* = \frac{1}{\lambda} = \frac{1}{n}\sum_i\sabs{\matW_i}
$$

\end{document}